\documentclass[review,12pt]{elsarticle}

\usepackage{amsmath,amsfonts,amssymb}
\usepackage[ruled, linesnumbered]{algorithm2e}
\usepackage{array}
\usepackage[caption=false,font=normalsize,labelfont=sf,textfont=sf]{subfig}
\usepackage{textcomp}
\usepackage{stfloats}
\usepackage{url}
\usepackage{verbatim}
\usepackage{graphicx}
\usepackage[table]{xcolor}
\usepackage{siunitx}
\usepackage{multirow}
\usepackage{booktabs}
\usepackage{adjustbox}
\usepackage{setspace}
\usepackage[hidelinks]{hyperref}
\usepackage{xcolor}

\newcommand{\cellbold}{\cellcolor{gray!50}}
\journal{}

\begin{document}

\begin{frontmatter}

\title{A Comparison-Relationship-Surrogate Evolutionary Algorithm for Multi-Objective Optimization}
\author{Christopher M. Pierce\corref{cor1}\fnref{uchicago}}
\author{Young-Kee Kim\fnref{uchicago}}
\author{Ivan Bazarov\fnref{cornell}}

\affiliation[uchicago]{organization={University of Chicago},
	addressline={5640 S Ellis Ave.}, 
	city={Chicago},
	postcode={60637}, 
	state={IL},
	country={USA}}
\cortext[cor1]{cmpierce@uchicago.edu}

\affiliation[cornell]{organization={Cornell University},
	addressline={245 Feeney Way}, 
	city={Ithaca},
	postcode={14853}, 
	state={NY},
	country={USA}}

\begin{abstract}
	Evolutionary algorithms often struggle to find well converged (e.g small inverted generational distance on test problems) solutions to multi-objective optimization problems on a limited budget of function evaluations (here, a few hundred).
	The family of surrogate-assisted evolutionary algorithms (SAEAs) offers a potential solution to this shortcoming through the use of data driven models which augment evaluations of the objective functions.
	A surrogate model which has shown promise in single-objective optimization is to predict the ``comparison relationship'' between pairs of solutions (i.e. who's objective function is smaller).
	In this paper, we investigate the performance of this model on multi-objective optimization problems.
	First, we propose a new algorithm ``CRSEA'' which uses the comparison-relationship model.
	Numerical experiments are then performed with the DTLZ and WFG test suites plus a real-world problem from the field of accelerator physics.
	We find that CRSEA finds better converged solutions than the tested SAEAs on many of the \emph{medium-scale, biobjective} problems chosen from the WFG suite suggesting the ``comparison-relationship surrogate'' as a promising tool for improving the efficiency of multi-objective optimization algorithms.
\end{abstract}

\begin{keyword}
Surrogate-Assisted Evolutionary Algorithm (SAEA), Expensive Optimization, Medium-Scale Multi-objective Optimization, Comparison-Relationship Surrogate, Pairwise Comparison.
\end{keyword}

\end{frontmatter}

\section{Introduction}
\label{sec:intro}
Across a diversity of fields, the users of optimization algorithms often face problems that involve multiple expensive-to-evaluate objective functions~\cite{john_antenna_2009, grosdidier_eadock_2007, castillo_tapia_applications_2007, bazarov_multivariate_2005, lian_multi-objective_2005}.
The expense is usually the time taken to numerically evaluate the objectives.
However, more and more, it can also be a monetary cost associated with a real-world measurement~\cite{roussel_multiobjective_2021, knowles_closed-loop_2009,clayton_automated_2020}.
For users operating on a budget, this requires the optimization task to be completed using only a small fixed number of function evaluations.
In the context of this paper, we take that number to be a few hundred in total.

Solving multi-objective optimization problems with potentially black-box functions is commonly achieved using a genetic algorithm.
Unfortunately, these methods are considered by many to be sample inefficient and can require a large number of function evaluations (extending to the tens of thousands) to produce a reasonable set of solutions~\cite{jin_surrogate-assisted_2011-1, veldhuizen_multiobjective_2000}.
When faced with a limited budget, this inefficiency may force users to sacrifice the quality of their results.
Additionally, some problems may be so expensive that they are not solvable, even approximately, within the user's limits.
This points to a need for new efficient multi-objective genetic algorithms that produce well-converged solutions using only a small number of function evaluations.

A popular approach to increasing the sample efficiency of multi-objective genetic algorithms is to introduce a surrogate model: a fast-to-evaluate model of the objectives based on already evaluated data.
These surrogate models (also called meta-models) can act as a stand-in for the expensive objective function evaluations in existing optimization algorithms.
For example, in K-RVEA~\cite{chugh_surrogate-assisted_2018}, a model of the objective functions is generated from an initially evaluated set of solutions and the optimization is performed against the model using a reference-vector-based evolutionary algorithm.
The model is then periodically updated with real calls to the objective functions for candidate solutions.
For a review of surrogate-assisted evolutionary algorithms (SAEAs), readers are pointed to~\cite{jin_surrogate-assisted_2011-1,jin_comprehensive_2005,deb_surrogate_2021}.

Instead of directly modeling the objective functions' values, a possible alternative is to build a surrogate of their comparison relationships.
That is, a model of the boolean function,
\begin{equation}
	c_i(\vec{x}_1, \vec{x}_2) := f_i(\vec{x}_1) < f_i(\vec{x}_2),
	\label{eq:comp-relationship}
\end{equation}
where $\vec{x}_1$ and $\vec{x}_2$ are the decision variables for the two candidate solutions being compared, and $f_i$ are the objective functions.
The problem of building an accurate surrogate model has now become one of classification instead of regression and the domination relationships can be inferred from the comparisons between objective functions.
In this work, we explore integrating this type of model into a multi-objective genetic algorithm to improve its efficiency in solving expensive optimization problems.

The main contributions of this work can be summarized as the following.
\begin{enumerate}
	\item A surrogate model of the comparison relationships between the objective functions of two solutions is proposed for solving expensive multi-objective optimization problems. This model is implemented using a neural network with an architecture designed to reflect the symmetry properties of the comparison operator.
	\item A method of reconstructing the ranking of solutions from their approximate comparisons (which may not satisfy the transitive property) is explored.
	\item The model and its associated improvements are integrated into a new SAEA (called ``CRSEA'') which is then compared experimentally on a set of benchmark problems to contemporary algorithms for solving expensive multi-objective optimization problems on a budget.
\end{enumerate}

The rest of this article is dedicated to exploring the use of a comparison-relationship-surrogate model to improve the sample efficiency of multi-objective genetic algorithms.
In Sec.~\ref{sec:related-literature}, we review the existing literature surrounding surrogate-assisted evolutionary algorithms (SAEAs) which are powered with classification-based models and the state of comparison-relationship models for optimization.
We describe the new SAEA that uses a model of the comparison relationships in Sec.~\ref{sec:algorithm-description}.
The construction of the model using a neural network is also described and architectural / training details related to the symmetries of the comparison operator are discussed.
In Sec.~\ref{sec:empirical-studies}, we report on numerical experiments that compare the new algorithm to a diverse set of existing SAEAs.
These experiments are performed on several well-known suites of test problems.
Results from the algorithm on an example real-world optimization problem are also given in Sec.~\ref{sec:performance-real-world-problem}.
We conclude the paper in Sec.~\ref{sec:conclusion} and give suggestions for future areas of study related to the comparison-relationship-surrogate optimization algorithm.

\section{Related Work}
\label{sec:related-literature}

\subsection{Regression-Based SAEAs}
One approach to building a data-driven model of the already evaluated solutions is to treat the situation as a regression problem.
We can attempt to predict the values of the objective functions or metrics related to them using the decision variables.

The algorithm ParEGO~\cite{knowles_parego_2006}, for example, uses a Gaussian process model of a scalarization function of the objectives.
New points are chosen to maximize the expected improvement of the solution in a manner similar to the kriging method~\cite{oliver_kriging_1990}.
SMS-EGO extended the idea to maximize improvement in the hypervolume (S-metric) instead~\cite{ponweiser_multiobjective_2008}.
In K-RVEA, the regression model is used directly in place of the expensive objective function evaluations in a reference-vector guided evolutionary algorithm~\cite{chugh_surrogate-assisted_2018}.
Multi-objective optimization problems are first decomposed into single objective sub-problems in the algorithm MOEA/D-EGO~\cite{zhang_expensive_2010}.
A Gaussian process is used to model each of the scalar sub-problems which are then solved simultaneously to suggest new candidate solutions to evaluate.

\subsection{Classification-Based SAEAs}
Surrogate assisted evolutionary algorithms driven by classification models have shown some success in improving the efficiency with which expensive optimization problems can be solved on a budget.
In early work, the algorithm Pareto-SVM~\cite{loshchilov_mono_2010} was introduced which uses a support vector machine to classify regions of decision space into those which map to points in objective space that are part of the dominated set, the current Pareto front, or unexplored regions.
A variant of the algorithm~\cite{deb_dominance-based_2010} followed shortly after and introduced a new classification model that accepts pairs of solutions and approximates their domination relationship directly.
The themes of classifying either regions of decision space or the relationships between pairs of solutions can be seen throughout later work and we use them to help organize the remainder of this section.

\subsubsection{Decision Space Classification}
We begin by taking a look at some of the algorithms that use classification models that accept a single point in decision space as their input.
In CPS-MOEA~\cite{zhang_classification_2015}, the solutions in the training set are labeled into equally sized positive and negative classes using a combination of non-domination rank and crowding distance.
A k-nearest neighbor classifier is then trained on the data and used to pre-filter candidate solutions before evaluation.
The algorithm MOEA/D-SVM~\cite{lin_decomposition_2016} explored using a scalarization function on the objectives to split the population into positive and negative classes.
In one proposed algorithm~\cite{zhang_preselection_2018}, a ``one-class classification model'' avoided the problem of splitting the data by only considering positive training examples.
CSA-MOEA~\cite{li_classification_2022} explored using a decision tree to sort regions of decision space into those which were and were not promising to find candidates in.
New infill strategies were explored for choosing which candidates to evaluate.
Multiple surrogate models were applied to solve ``local'' problems in a decomposition-based optimization algorithm to mitigate overfitting in MCEA/D~\cite{sonoda_multiple_2022}.
The algorithm CR-SAEA~\cite{geng_classification_2024} augments the multiple local classification models approach with a global regression model to help guide its decomposition-based genetic algorithm.

\subsubsection{Relationship-Based Surrogates}
Modeling the relationships between solutions (e.g. domination) is also an important theme in multi-objective optimization algorithms aided by classification models.
The concept of modeling the Pareto-dominance relationship was explored further in some work~\cite{bandaru_performance_2014} which compared ten off-the-shelf classification models for use in this task.
The algorithm CSEA~\cite{pan_classification-based_2019} was introduced which uses a surrogate model to estimate domination between candidate solutions and selected already evaluated reference solutions.
The algorithm also uses a measure of uncertainty in the predictions to choose which candidates to evaluate.
Two surrogate models: one of the Pareto-domination relationships and a second of the $\theta$-domination relationships between candidates are combined in the algorithm $\theta$-DEA-DP~\cite{yuan_expensive_2022}.
In REMO~\cite{hao_expensive_2022}, the relationships between candidate solutions are modeled with a neural network classifier trained on pairs of solutions which are grouped into equal sized sets based on domination.
A related algorithm, named CREMO~\cite{hao_relation_2023}, also explored relationship learning using a convolutional neural network on the decision variables.
In FAEA-SDR~\cite{shen_expensive_2023}, the strengthened dominance relationship between solutions is modeled instead.
Pairwise comparisons of the scalarized objective functions are learned and used in the optimization algorithm PC-SAEA~\cite{tian_pairwise_2023}.
The concept of domination modeling has also been applied to neural architecture search with promising results~\cite{ma_pareto-wise_2024}.

\subsubsection{Comparison-Operator Models}
Some work has been put into modeling the comparison relationships between the objective functions directly.
That is, the value of the comparison operator evaluated on the objective functions between pairs of solutions.
The use of these models has remained restricted to the case of single-objective optimization or constraint satisfaction problems.
One such study~\cite{runarsson_ordinal_2006} showed an algorithm that ranked solutions using a support vector machine could outperform evolutionary algorithms aided with regression models on some benchmark problems.
Binary relationship learning for the pre-selection of solutions was investigated in Hao et al~\cite{hao_binary_2020}.
Multiple classification models were compared for the purpose of ranking solutions based on their comparison relationships in single-objective optimization.
A wide variety of pairwise classification models and regression models were compared against each other on benchmark problems for single-objective optimization in~\cite{naharro_comparative_2022} with some evidence for the success of the classification approach.
A specially designed symmetric neural network architecture was created for the task of modeling the comparison operator between multiple objectives in~\cite{hakhamaneshi_bagnet:_2019} and applied to solve constraint satisfaction problems in integrated circuit design.
While these results demonstrate the potential of a comparison-based model, they did not address the unique challenges of integrating one with a multi-objective optimization algorithm.
This includes integrating the comparison predictions across all objectives, diversity preservation in higher dimensions, and dealing with predictions that may not satisfy important symmetries of the comparison operator which are relied on in (for example) domination sorting.

\begin{algorithm}
	\setstretch{1}
	\caption{CRSEA Optimization Algorithm}
	\label{alg:crsea}
	\DontPrintSemicolon
	\KwIn{$\text{FE}_{\max}$, the maximum number of function evaluations; $N$, the population size; $w_{\text{max}}$, the number of generations executed before updating the model; $\mu$, the number of expensive function evaluations performed during a model update.}
	\KwOut{$P$, the final population}
	
	\BlankLine
	\tcc{Initialize population and model}
	$P_t \gets$ initial set of $N_i$ solutions, Latin hypercube\;
	$P \gets$ $N$ solutions selected from $P_t$ as in NSGA-II\;
	$\theta \gets$ model parameters from initial training on $P_t$\;
	FE $\gets N_i$ \tcp*{Track function evaluations}
	
	\BlankLine
	\While{$\text{FE} \leq \text{FE}_{\max}$}{
		\BlankLine
		\tcc{Run NSGA-II using the model}
		$w \gets 1$ \tcp*{"Model generation" count}
		$P_m \gets P$ \tcp*{Model population}
		$\text{fit} \gets$ fitness calculated with nondominated sorting\;
		
		\BlankLine
		\While{$w \leq w_{\text{max}}$}{
			$c_m \gets$ get $N$ children, use ``fit'' in tournament\;
			$P_m \gets P_m \cup c_m$\;
			$P_m \gets$ select $N$ solutions in $P_m$ using model\;
			$\text{fit} \gets$ calculate solutions' fitness with model\;
			$w \gets w + 1$\;
		}
		
		\BlankLine
		\tcc{Evaluate Solutions }
		$c \gets$ evaluate $\mu$ random candidate solutions from $P_m$\;
		$\text{FE} \gets \text{FE} + \mu$\;
		
		\BlankLine
		\tcc{Update model}
		$P_t \gets P_t \cup c$ \tcp*{Add to training set}
		$\theta \gets$ train model on $P_t$\;
		
		\BlankLine
		\tcc{Select solutions based on fitness}
		$P \gets P \cup c$\;
		$P \gets$ select $N$ solutions from $P$ as in NSGA-II\;
	}
\end{algorithm}

\begin{figure}[!t]
	\centering
	\includegraphics[width=0.7\linewidth]{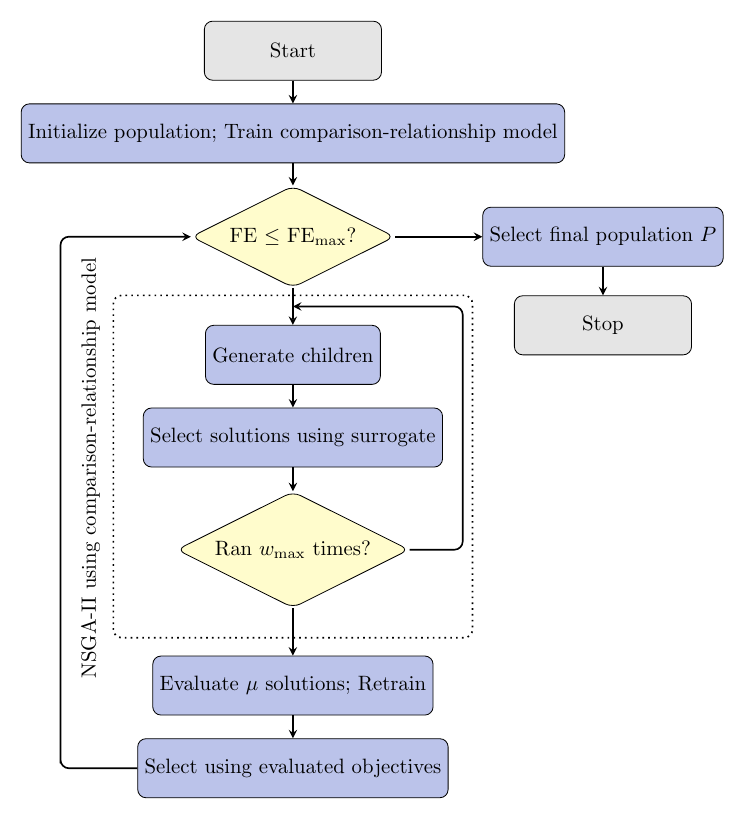}%
	\caption{
		A diagram of the how the CRSEA algorithm (Alg.~\ref{alg:crsea}) works.
		An initial population is generated and the comparison-relationship model trained.
		Then, the comparison-relationship model is used in the place of the objective functions in an evolutionary algorithm for some number of generations ($w_{\text{max}}$).
		Several ($\mu$) randomly chosen solutions from the resulting candidate solutions are evaluated and used to retrain the model.
		The process is repeated until the budget of function evaluations ($\text{FE}_{max}$) is exhausted.
	}
	\label{fig:algorithm}
\end{figure}

\subsection{Motivation}
Unlike a model of the objective functions themselves, models of the comparison relationship are invariant under all monotonic transformations of the objective functions.
Some point to the invariance properties of evolutionary algorithms as one reason for their observed robustness on a wide variety of problems~\cite{schaefer_comparison-based_2010, gelly_comparison-based_2007, auger_experimental_2009}. 
When building a surrogate model for an evolutionary algorithm, it is therefore natural to consider one such as a comparison-based surrogate that reflects the same symmetries.

In addition to the invariance properties of the model, a classification problem is always ``simpler'' than the equivalent regression problem in the sense that it is solved in the process of solving the regression problem, but does not solve the regression problem itself.
That is, a model of the objective function values also gives the classifications whether they predict domination, comparisons, or something else.
However, knowledge of what class a solution lies in is not enough information to determine the values of the objective functions.
It may be that this ``simpler'' problem is easier to solve.

The task of learning the comparison relationships between the objectives also forms a naturally balanced classification problem.
In a model of the domination relationships, for example, as the algorithm drifts closer to the Pareto front most new examples are dominated by the existing population.
The comparisons between objective functions, on the other hand, can always be manipulated to have an equal number of positive and negative examples by swapping the order of the two solutions being compared.

With these arguments and the previously discussed success of comparison-relationship models in solving single-objective optimization problems, we are motivated to use a similar model to drive a genetic algorithm in solving the multi-objective case.

\section{A Comparison-Relationship-Surrogate multi-objective Genetic Algorithm}
\label{sec:algorithm-description}
With an accurate enough model of the comparison relationships (Eq.~\ref{eq:comp-relationship}), the expense of evaluating the objective functions in an evolutionary algorithm could be replaced completely with a cheaper call to the model.
This approach for integrating information from the model into an optimization algorithm has seen success in other SAEAs (such as in K-RVEA~\cite{chugh_surrogate-assisted_2018}) which motivates us to handle the comparison-relationship model in a similar manner.

With this inspiration, we propose the algorithm described in Alg.~\ref{alg:crsea}.
The model is first initialized using training points that are drawn from the decision space with Latin hypercube sampling~\cite{mckay_comparison_2000} and evaluated using the real objective functions.
Candidate solutions are found from the model using an evolutionary algorithm.
To combat the model's loss of accuracy when it is evaluated far from the points it was trained on, the model is periodically updated using a small number of candidate solutions for which the real objective function values are computed.
This process (solving the optimization problem and updating the model) is repeated until the budget of function evaluations is exhausted.
A diagram of the algorithm is shown in Fig.~\ref{fig:algorithm}.

The main difference between the new algorithm and existing SAEAs is the use of the comparison-relationship model which directly models the comparison relationships between each objective of every pair of solutions.
As discussed in the motivation section, a benefit of this model is that it has the same invariance under monotonic transformations of the objective functions as the genetic algorithms it is being integrated with.
It also forms a naturally balanced classification problem solving some difficulties in other classification-based SAEAs.
On the other hand, diversity preservation with a model that doesn't have information about the distance between solutions may be challenging.
In the following subsections, we explain in detail the integration of the comparison-relationship model into a genetic algorithm.

\subsection{Integrating the Comparison-Relationship Model with an Evolutionary Algorithm}
\label{sec:calculating-approximate-pf}
The major step in the proposed multi-objective optimization algorithm is generating candidate solutions using only the model.
Because the surrogate model can be evaluated cheaply, this step may be performed with a conventional multi-objective genetic algorithm.
The lack of information about the values of the objectives from the comparison-relationship model, however, imposes some limits on which types of algorithms can be run on it.
As an example, algorithms that use reference vectors for diversity preservation require some knowledge of the distance or angle between solutions and are ruled out.

With this restriction in mind, we propose finding optimal solutions from the model using NSGA-II~\cite{deb_fast_2002}.
Briefly summarizing the algorithm, in each generation, a set of children is produced with mutation, crossover, and selection operators (in this paper, polynomial mutation, simulated binary crossover, and tournament selection~\cite{deb_multi-objective_2001}).
After evaluating the objective functions, the population is sorted by increasing rank of non-domination.
Within those ranks, the ``crowding distance'' metric is used to break ties in order to promote diversity.
Solutions are selected from the newly sorted list with the smallest rank, largest crowding distance solutions being the most fit.
We suggest running the algorithm using only information from the surrogate model.

Although NSGA-II is well suited for use with the comparison-relationship model, there are still a few challenges that come with it.
Each is described in detail in the following subsections and we propose ways of mitigating them.
First, we find that approximate models of the comparison relationships which do not satisfy the transitive properties must have their results corrected before use in an optimization algorithm.
We then discuss the prediction of domination from a model that only predicts strict inequality between solutions.
We conclude by talking about the crowding distance metric and the difficulty encountered in using it during selection with a model that only predicts comparisons.

\subsubsection{Violation of the Transitive Property}
The comparison operator satisfies the transitive property.
For the objective functions (single objective for simplicity) of three solutions, $f_1$, $f_2$, $f_3$, if $f_1 < f_2$ and $f_2 < f_3$ then $f_1 < f_3$ must be true.
Approximate models of the comparison operator may violate the transitive property.
This can lead to paradoxical situations such as the model predicting a solution 1 dominates solution 2, solution 2 dominates solution 3, but solution 3 is not dominated by solution 1.
Such nonsensical results can wreak havoc on algorithms such as the fast non-dominated sorting algorithm from Ref.~\cite{deb_fast_2002} used in NSGA-II.
To integrate a comparison-relationship model into the genetic algorithm, we must find a way of avoiding these situations.

We propose ``cleaning'' the predictions of the comparison model so that they respect the transitive property.
For all solutions $x_i$ in $\mathbb{P}$, the population being sorted, we can calculate the noisy predictions from the comparison model.
The predictions take the form of a probability that the comparison is true, $\mathcal{P}(f(x_i) < f(x_j))$, where $f(x)$ is the objective function (a single objective for simplicity).
To clean the predictions, we calculate the value,
\begin{equation}
	\tau_i = \frac{1}{|\mathbb{P}|}\sum_{j\in\mathbb{P}}\mathcal{P}(f(x_i) < f(x_j)),
	\label{eq:noisy-sorting}
\end{equation}
from Ref.~\cite{shah_simple_2018}.
We take the value of $\mathcal{P}(f(x_i) < f(x_i))$ as zero.
The quantity $\tau_i$ can be interpreted as the probability of the objective function of solution $i$ being smaller than that of a solution chosen at random from $\mathbb{P}$.
The comparison predictions defined by $f(x_i) < f(x_j) := \tau_i > \tau_j$  now satisfy the transitive property.
This cleaning procedure is repeated for all objective functions and the domination relationships can be calculated from there, allowing the fast non-dominated sorting algorithm to be run using the comparison model.

\subsection{Equality and the Comparison-Relationship Model}

Assuming a minimization problem, one solution dominates another if all of its objectives are less than or equal to the other's and at least one of its objectives is strictly less than the corresponding objective in the other solution.
Unfortunately, the comparison-relationship model seems to only predict the case of strict inequality.
This prevents us from predicting domination in the cases where some of the objective for two solutions are equal.
However, equality happens exactly at the point in between the region where $f(x_1) < f(x_2)$ and the region where $f(x_2) < f(x_1)$.
In terms of the model, these are exactly the points which lie along the classifier's decision boundary!
This allows us to define domination as predicted from the model as $\tau_i <= \tau_j$ for all objectives and $\tau_i < \tau_j$ for at least one using our ``cleaned'' predictions as described in the previous section.

\subsubsection{Crowding Distance}
With the ability to clean the model's predictions from violations of the transitive property, we can now use it to predict the non-domination rank of solutions.
In NSGA-II, ``crowding distance'' is then used to break ties of non-domination rank for the purpose of selection.
This acts as a form of diversity preservation.

To calculate crowding distance, each solution is assigned a value of zero to start with.
For each objective, the solutions are sorted by the value of their objective function.
The extreme solutions (highest and lowest value) are assigned a crowding distance of infinity.
Every other solution has its crowding distance incremented by the distance between its two nearest neighbors for this objective normalized by full range spanned by all solutions (distance between the biggest and smallest values).

The comparison-relationship model, of course, does not have information on the distance between neighbors.
Instead we suggest using the values of $\tau$ from Eq.~\ref{eq:noisy-sorting} to predict the solutions with extreme values for the objective functions (smallest and largest).
These solutions are preferred when breaking ties of non-domination rank.
This should achieve a similar effect to setting their crowding distance to infinity in NSGA-II, but we do expect worse diversity preservation since the surrogate model cannot predict the distance between solutions.

With these two adjustments (cleaning the predictions to satisfy the transitive property, and breaking ties with predictions of extreme solutions instead of crowding distance) the comparison-relationship model may be used in an NSGA-II-like algorithm to perform selection.

\subsection{Modeling the Comparison Relationship}
\label{sec:model}
One way to approach the task of modeling the comparison relationships between objectives is by considering it as multiple binary classification problems.
The training data here is the set of pairs of solutions in the evaluated population with the labels being the $m$ comparisons of objective functions ($f_i(x_1) < f_i(x_2)$) for each example.
We are interested in building a model from the data to predict the comparisons for the unseen pairs of solutions.
Due to their expressive power and success in solving problems with high dimensional inputs, we used a deep neural network to approximate the comparison relationship.

\begin{figure*}[!t]
	\centering
	\adjustbox{width=\textwidth + 4cm, center}{
		\includegraphics[width=\linewidth]{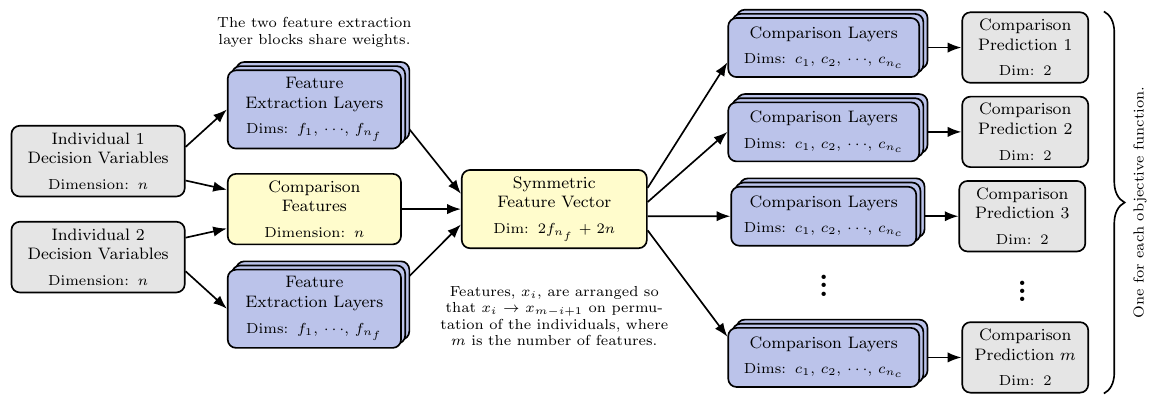}%
	}
	\caption{
		A schematic of the comparison surrogate model. Decision variables of the two solutions being compared are fed into feature extraction networks made of densely connected layers with shared weights.
		Additional features in the form of the comparisons $x_{1,i} < x_{2,i}$ are generated.
		All features are combined into a symmetric vector such that a permutation of the two input solutions will cause the feature vector to reverse itself ($f_i\to f_{n-i+1}$ for a vector $f$ with $n$ elements).
		These features are fed into comparison networks (one for each objective) which use special symmetric dense layers to ensure the correct behavior under permutation of the inputs.
	}
	\label{fig:nn_architecture}
\end{figure*}

The comparison operator has the symmetry,
\begin{equation}
	\label{eq:symmetry}
	f_i(x_1) < f_i(x_2) \implies f_i(x_1) \ngtr f_i(x_2),
\end{equation} 
which is not guaranteed in a naive implementation using a neural network.
We fix this by using the architecture proposed in~\cite{hakhamaneshi_bagnet:_2019}.
A sketch of the architecture is shown in Fig.~\ref{fig:nn_architecture} and we briefly describe it here.

Considering a network that outputs predictions with the one-hot encoding scheme, the requirement in Eq.~\ref{eq:symmetry} is equivalent to requiring that the output vector of the network is reversed on permutation of the two inputs.
To achieve this, the decision variables for the two solutions being compared are each first fed into a feature extraction network that consists of conventional dense layers with ReLU activation and has shared weights and biases between the two inputs.
The output features of the two solutions are concatenated into a single feature vector with the elements from the second network being reversed so that permuting the solutions is equivalent to reversing the full feature vector.
That is,
\begin{equation}
	(\vec{f}_a, \vec{f}_b) \to [f_{a,1}, f_{a,2}, \cdots, f_{a,n}, f_{b,n}, \cdots, f_{b,2}, f_{b,1}],
\end{equation}
where $\vec{f}_a$ and $\vec{f}_b$ are the outputs of the two feature extraction networks and $n$ is the size of the final layer in the feature extraction network.

The complete feature vector is then fed into a separate ``comparison network'' for each objective.
Because the requirement from Eq.~\ref{eq:symmetry} is already satisfied by the feature vector, to ensure it for the whole network we only need to preserve the symmetry in the comparison layers,
\begin{align}
	y_i = f\left(w_{i,j} x_j + b_j\right),
\end{align}
where $f(x)$ is the activation function, $w_{i,j}$ and $b_i$ are the weights and biases, and we are using Einstein summation notation.
The layers will preserve our required symmetry as long as they satisfy the conditions,
\begin{align}
	w_{i,j} = w_{m-i+1,n-j+1}, && b_{i} = b_{n-i+1},
\end{align}
where $m$ and $n$ are the dimensions of the weights matrix.
In our work, the ReLU non-linearity was used for these comparison layers except for the output layer which used the softmax activation function.
All weights are initialized to be uniformly distributed in the way described in Ref.~\cite{glorot_understanding_2010}.
With these constraints, the full model will only produce predictions with the symmetry described in Eq.~\ref{eq:symmetry} on permutation of the inputs.

\begin{figure}[!b]
	\centering
	\includegraphics{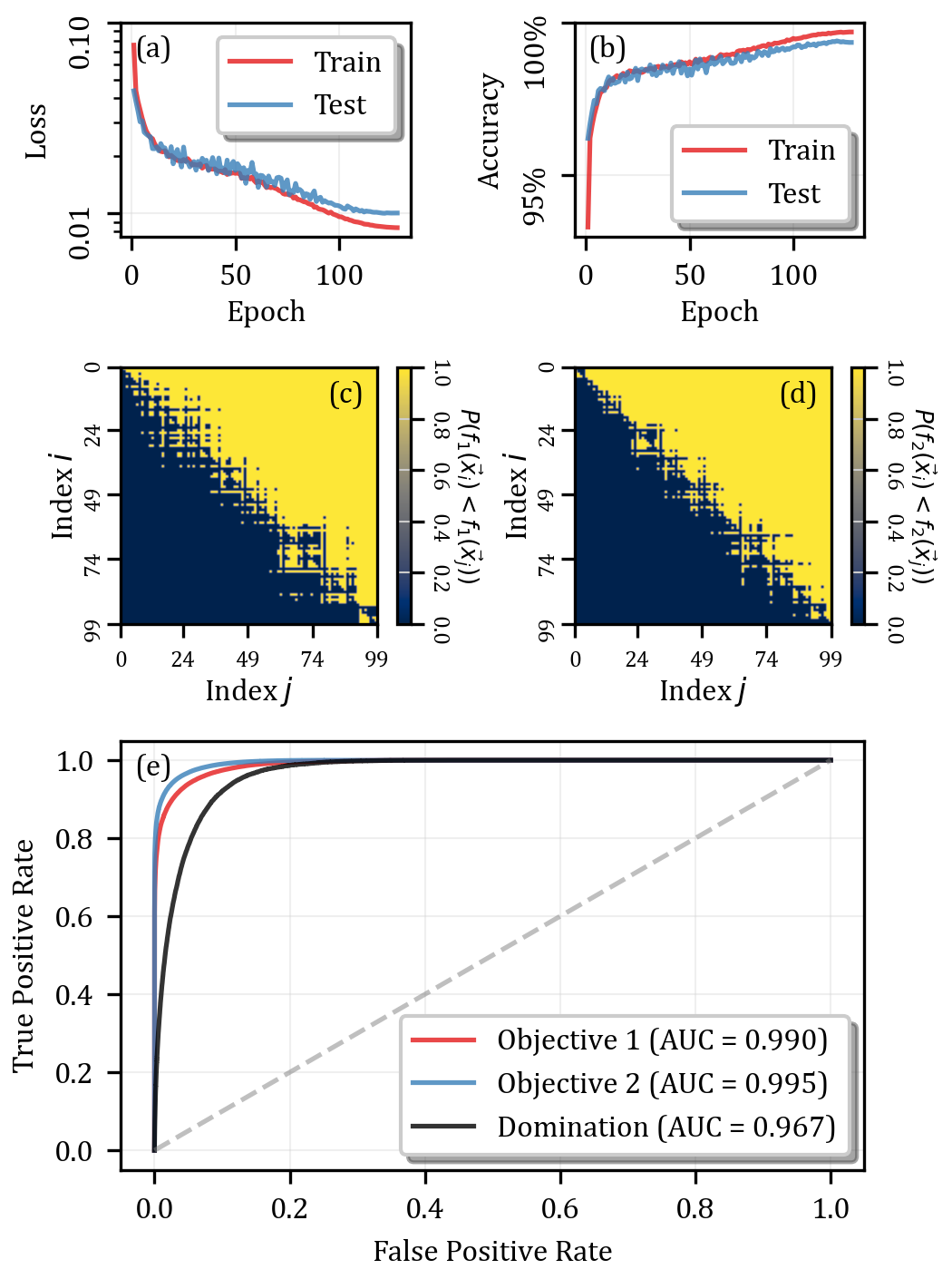}%
	\caption{
		An example of the comparison model trained on data from the WFG8 ($n=16$) problem. (a) loss function on the train and test set during training; (b) the accuracy during training; (c) and (d) predicted comparisons for both objectives on never-before-seen samples where solutions are indexed in sorted order for the objective being compared (i.e. predictions from a perfect classifier would show up as a solid yellow upper triangle); (e) the receiver operating characteristic curve for each model output as well as domination predictions.
	}
	\label{fig:model-comparisons}
\end{figure}

Besides features learned from the training data, additional features may be added manually as long as they are introduced to the final feature vector in a way that preserves its symmetry.
For example, we suggest adding the features $x_{1,i} < x_{2,i}$ ($1\leq i \leq n$) to allow the comparison networks to learn from the raw comparisons themselves.
These are added to the middle of the feature vector along with a reversed and inverted copy to ensure the correct behavior under permutation of the inputs.
These extra features could be useful in several practical applications where one of the objectives may be a monotonic function of a single decision variable.

Training data can be generated from a set of decision variables and the corresponding values of the objective functions.
For each pair of solutions in the training set (discounting duplicates like $a<b$ and $b<a$), the comparisons between objective functions are evaluated.
An error term is calculated from the model for each objective using the binary cross entropy loss function.
The total loss is then optimized using mini-batch gradient descent with the AdamW optimizer~\cite{loshchilov_decoupled_2019} and cosine annealing with warm restarts~\cite{loshchilov_sgdr_2017} where the learning rate scheduler is ``restarted'' at the beginning of every retraining step in the algorithm.
When the model is used in a multi-objective optimization algorithm, we retain all evaluated solutions for use in the training set and retrain the model at each generation of the optimizer (as in Alg.~\ref{alg:crsea}).

\begin{figure}[h!]
	\centering
	\includegraphics{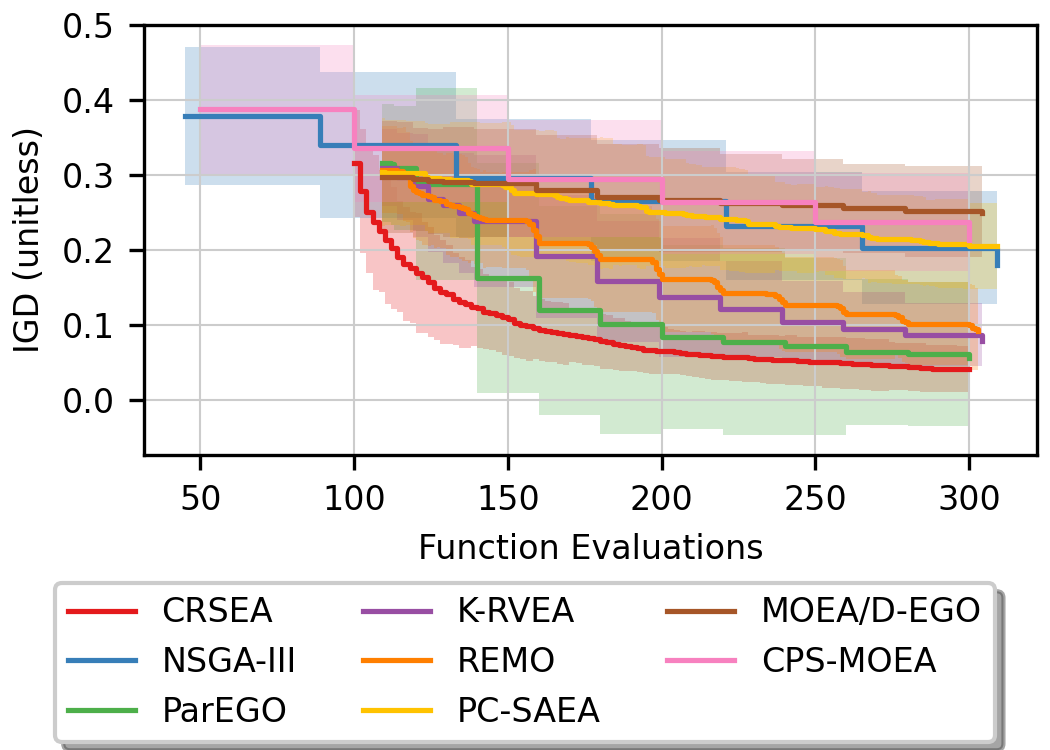}%
	\caption{
		Mean value of the IGD metric and 95\% prediction interval for non-dominated solutions to DTLZ5 ($m=3$, $n=10$).
	}
	\label{fig:dtlz-performance}
\end{figure}

\begin{figure}[h!]
	\centering
	\includegraphics{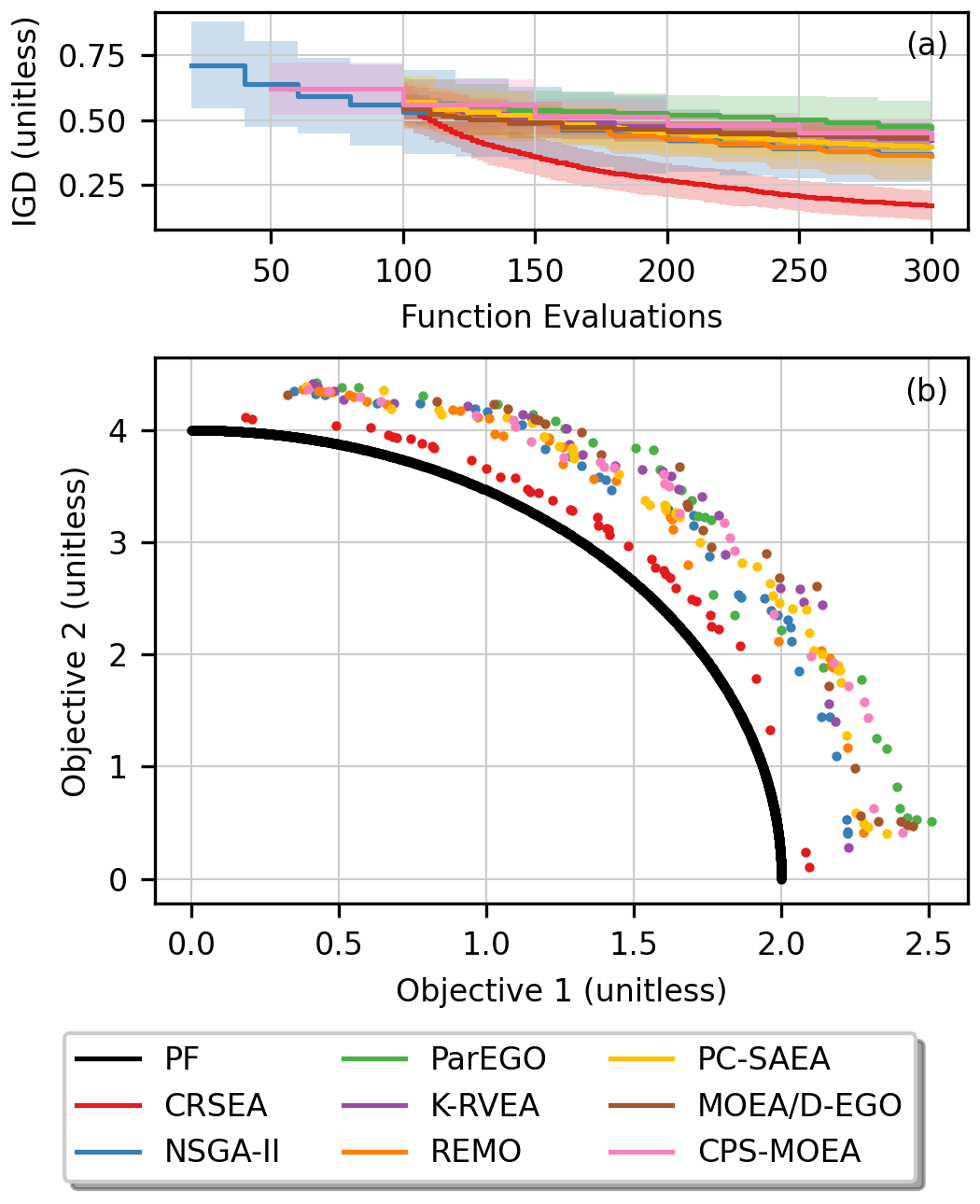}%
	\caption{
		Results from the experiments performed in Tab.~\ref{tab:high-dimension-results} for the WFG7 (n=16) test problem : (a) Mean value of the IGD metric and 95\% prediction interval for non-dominated solutions; (b) Solutions after 300 function evaluations from the optimization with the median value of the IGD metric for each algorithm. The analytical Pareto front (labeled ``PF'') is plotted in black.}
	\label{fig:wfg-combined}
\end{figure}
An example of using the comparison-relationship-surrogate model is shown in Fig.~\ref{fig:model-comparisons}.
A training set of 300 solutions was generated for the problem WFG8 ($n=16$)~\cite{huband_review_2006} and the model was trained on it for 128 epochs.
The resulting learning curves are shown in subfigures (a) and (b) using an 80\%-20\% train-test split of the labeled comparison data.
Predicted comparisons were then evaluated for every pair in a second set of 128 randomly generated solutions.
The results are plotted in subfigures (c) and (d) where the solutions are indexed according to their position in the array sorted by the objective being evaluated.
This representation helps visualize when and where errors occur in the comparisons.
For example, many of the out-of-place elements are close to the diagonal meaning that errors tend to be between solutions that are similar in value for that objective.
In subfigure (e) we show the receiver operating characteristic curve and list the area under curve metric for the comparison models as well as for the resulting domination classifier.
As expected, the domination predictions are degraded as a correct prediction requires that both underlying comparisons are predicted correctly.

\section{Empirical Studies of Performance}
\label{sec:empirical-studies}
In this section, we evaluate the performance of the comparison-relationship-surrogate optimizer through a set of numerical experiments.
In each experiment, the new optimizer, as well as a set of contemporary algorithms, are used to solve test problems on a fixed budget of 300 function evaluations.
The performance of the algorithms is compared based on the inverted generational distance~\cite{coello_solving_2005} of the returned solutions to the analytical Pareto fronts.
Due to the stochastic nature of evolutionary optimization, these comparisons are carried out using the Wilcoxon rank sum statistical test at a significance level of 95\% with a sample size of 32.

Three different classes of algorithms were chosen to compare against: conventional genetic algorithms, regression-based surrogate-assisted algorithms, and classification-based surrogate-assisted algorithms.
From the first class, NSGA-II~\cite{deb_fast_2002} was chosen for test problems with two objectives.
This algorithm was selected for its popularity in applications as well as the fact that the new comparison-surrogate-assisted algorithm is strongly influenced by it.
For the problems with more than two objectives, NSGA-III~\cite{deb_evolutionary_2014} was chosen as the conventional optimizer due to its improved performance on many-objective problems.
From the category of regression-based surrogate-assisted optimizers, we chose ParEGO~\cite{knowles_parego_2006}, K-RVEA~\cite{chugh_surrogate-assisted_2018}, and MOEA/D-EGO~\cite{zhang_expensive_2010} for a combination of their performance on test problems and popularity in the literature.
Because the new algorithm being studied uses a classification-based surrogate model, we feel it is important to include similar algorithms in the numerical experiments.
For these purposes, we chose an example of an algorithm that classifies the regions of decision space, CPS-MOEA~\cite{zhang_classification_2015}, as well as two that use a pairwise model of the domination relationships, REMO~\cite{hao_expensive_2022} and PC-SAEA~\cite{tian_pairwise_2023}.

All algorithms except for the new comparison-relationship-surrogate-based optimizer were evaluated using their implementation in PlatEMO~\cite{tian_platemo_2017} version 4.11 with default settings for all parameters.
For algorithms that require a population size, it was set to 50 in all cases, except for NSGA-II where it was set to 20 following the recommendation of Ref.~\cite{knowles_parego_2006} and NSGA-III where it was set based on the number of objectives using the table in Ref.~\cite{pan_classification-based_2019}.
For surrogate-assisted algorithms that use an initial population size, it was set to $N_i = 11n-1$ as recommended for the Latin hypercube sampling technique~\cite{mckay_comparison_2000} and was capped at $N_i = 100$ for problems with more than ten decision variables to avoid immediately exhausting the budget of function evaluations.

\begin{table*}
	\caption{
		Statistical results for the IGD metric of 32 runs of each optimizer on the DTLZ problem set with ten decision variables. Cells that are among the best performing for that problem up to a 95\% significance level are highlighted.
	}
	\label{tab:many-objective-results}
    \adjustbox{width=\textwidth + 3cm, center}{
	\centering
	\vspace{3mm}
	\normalsize{
		\begin{tabular}{llllllllll}
	\toprule
	\multicolumn{1}{c}{\textbf{Problem }} &\multicolumn{1}{c}{\textbf{ m }}  &  \multicolumn{1}{c}{\textbf{ NSGA-III }}&  \multicolumn{1}{c}{\textbf{ ParEGO }}&  \multicolumn{1}{c}{\textbf{ K-RVEA }}&  \multicolumn{1}{c}{\textbf{ MOEA/D-EGO }}&  \multicolumn{1}{c}{\textbf{ CPS-MOEA }}&  \multicolumn{1}{c}{\textbf{ REMO }}&  \multicolumn{1}{c}{\textbf{ PC-SAEA }}&  \multicolumn{1}{c}{\textbf{ CRSEA }}\\
	\midrule
	\multirow{5}{*}{DTLZ1} & 3 & 9.09e+1 (1.57e+1) - & 6.73e+1 (5.57e+0) + & 9.01e+1 (2.27e+1) - & 8.43e+1 (1.62e+1) $\approx$ & 8.00e+1 (1.57e+1) $\approx$ & \cellbold 5.40e+1 (1.87e+1) + & 8.61e+1 (1.66e+1) $\approx$ & 8.11e+1 (1.65e+1) \\
	 & 4 & 6.04e+1 (1.62e+1) $\approx$ & 5.63e+1 (7.24e+0) + & 6.23e+1 (1.41e+1) $\approx$ & 6.21e+1 (1.23e+1) $\approx$ & 5.94e+1 (1.35e+1) $\approx$ & \cellbold 3.78e+1 (1.20e+1) + & 7.03e+1 (1.60e+1) - & 6.26e+1 (1.23e+1) \\
	 & 6 & 3.11e+1 (1.07e+1) + & 3.71e+1 (5.59e+0) + & 2.94e+1 (7.54e+0) + & 3.32e+1 (7.47e+0) + & 3.09e+1 (8.56e+0) + & \cellbold 1.53e+1 (4.56e+0) + & 2.76e+1 (1.12e+1) + & 4.21e+1 (7.12e+0) \\
	 & 8 & 1.08e+1 (5.16e+0) + & 1.55e+1 (4.15e+0) $\approx$ & 9.37e+0 (3.43e+0) + & 1.28e+1 (3.27e+0) + & 1.01e+1 (3.56e+0) + & \cellbold 4.55e+0 (2.28e+0) + & 9.16e+0 (3.17e+0) + & 1.53e+1 (5.47e+0) \\
	 & 10 & 5.22e-1 (2.29e-1) + & 4.58e-1 (8.84e-2) + & 3.68e-1 (1.07e-1) + & 4.61e-1 (1.62e-1) + & 3.99e-1 (8.58e-2) + & \cellbold 2.95e-1 (6.39e-2) + & 3.91e-1 (8.21e-2) + & 6.91e-1 (2.99e-1) \\
	\cline{1-10}
	\multirow{5}{*}{DTLZ2} & 3 & 2.93e-1 (2.69e-2) - & 1.60e-1 (3.75e-2) - & 1.73e-1 (4.10e-2) - & 3.54e-1 (2.27e-2) - & 2.86e-1 (2.10e-2) - & 2.09e-1 (2.56e-2) - & 2.80e-1 (2.39e-2) - & \cellbold 8.36e-2 (1.03e-2) \\
	 & 4 & 3.52e-1 (2.95e-2) - & 4.28e-1 (3.04e-2) - & 2.41e-1 (3.13e-2) - & 3.95e-1 (2.07e-2) - & 3.95e-1 (3.16e-2) - & 2.71e-1 (2.33e-2) - & 3.51e-1 (2.01e-2) - & \cellbold 2.17e-1 (2.44e-2) \\
	 & 6 & 4.83e-1 (3.55e-2) $\approx$ & 5.33e-1 (2.92e-2) - & \cellbold 3.37e-1 (2.38e-2) + & 4.81e-1 (1.72e-2) $\approx$ & 5.36e-1 (2.93e-2) - & 4.04e-1 (4.00e-2) + & 4.69e-1 (2.70e-2) + & 4.85e-1 (2.62e-2) \\
	 & 8 & 5.83e-1 (2.78e-2) - & 6.59e-1 (2.77e-2) - & \cellbold 4.67e-1 (3.70e-2) + & 5.18e-1 (1.50e-2) + & 5.98e-1 (2.90e-2) - & 5.13e-1 (4.67e-2) + & 5.71e-1 (2.07e-2) - & 5.31e-1 (1.87e-2) \\
	 & 10 & 6.71e-1 (2.99e-2) - & 7.00e-1 (2.87e-2) - & \cellbold 5.41e-1 (2.32e-2) $\approx$ & 5.69e-1 (3.21e-2) - & 6.55e-1 (3.95e-2) - & 6.12e-1 (4.50e-2) - & 6.72e-1 (2.91e-2) - & \cellbold 5.40e-1 (1.71e-2) \\
	\cline{1-10}
	\multirow{5}{*}{DTLZ3} & 3 & 2.65e+2 (6.68e+1) - & \cellbold 1.75e+2 (8.48e+0) + & 2.24e+2 (5.49e+1) $\approx$ & 2.03e+2 (3.32e+1) + & 2.13e+2 (3.41e+1) $\approx$ & \cellbold 1.63e+2 (4.38e+1) + & 2.38e+2 (4.10e+1) - & 2.17e+2 (2.71e+1) \\
	 & 4 & 1.82e+2 (5.01e+1) $\approx$ & 1.50e+2 (9.92e+0) + & 1.73e+2 (2.97e+1) $\approx$ & 1.66e+2 (2.30e+1) $\approx$ & 1.58e+2 (3.09e+1) + & \cellbold 1.24e+2 (2.88e+1) + & 2.01e+2 (3.98e+1) - & 1.72e+2 (2.00e+1) \\
	 & 6 & 9.47e+1 (3.70e+1) + & 9.64e+1 (1.23e+1) + & 8.25e+1 (2.29e+1) + & 9.53e+1 (1.59e+1) + & 9.32e+1 (2.06e+1) + & \cellbold 5.60e+1 (1.81e+1) + & 1.01e+2 (2.66e+1) + & 1.13e+2 (1.04e+1) \\
	 & 8 & 3.27e+1 (1.27e+1) + & 4.61e+1 (1.00e+1) $\approx$ & 2.80e+1 (1.07e+1) + & 3.62e+1 (1.23e+1) + & 3.41e+1 (1.33e+1) + & \cellbold 1.87e+1 (8.97e+0) + & 2.55e+1 (8.64e+0) + & 4.52e+1 (1.13e+1) \\
	 & 10 & 2.02e+0 (1.75e+0) + & 1.58e+0 (4.97e-1) + & 1.22e+0 (3.10e-1) + & 1.43e+0 (4.37e-1) + & 2.20e+0 (1.36e+0) $\approx$ & \cellbold 9.63e-1 (2.68e-1) + & 1.29e+0 (2.72e-1) + & 1.90e+0 (6.49e-1) \\
	\cline{1-10}
	\multirow{5}{*}{DTLZ4} & 3 & 7.34e-1 (1.92e-1) $\approx$ & 5.31e-1 (1.31e-1) + & 4.15e-1 (1.19e-1) + & 6.01e-1 (6.79e-2) + & 5.62e-1 (6.06e-2) + & \cellbold 2.62e-1 (1.27e-1) + & 5.26e-1 (8.64e-2) + & 7.26e-1 (1.71e-1) \\
	 & 4 & 7.36e-1 (1.60e-1) $\approx$ & 6.93e-1 (1.02e-1) $\approx$ & 5.12e-1 (9.17e-2) + & 6.55e-1 (5.16e-2) $\approx$ & 6.16e-1 (3.91e-2) + & \cellbold 3.22e-1 (7.95e-2) + & 6.01e-1 (5.30e-2) + & 7.06e-1 (1.96e-1) \\
	 & 6 & 7.74e-1 (9.65e-2) - & \cellbold 5.00e-1 (7.57e-2) + & 5.42e-1 (4.58e-2) + & 6.54e-1 (3.23e-2) - & 6.34e-1 (2.36e-2) - & \cellbold 5.10e-1 (4.82e-2) + & 7.06e-1 (3.64e-2) - & 6.05e-1 (5.73e-2) \\
	 & 8 & 7.54e-1 (6.06e-2) - & \cellbold 5.50e-1 (3.76e-2) + & \cellbold 5.62e-1 (3.80e-2) + & 6.29e-1 (1.83e-2) $\approx$ & 6.31e-1 (1.36e-2) $\approx$ & 5.92e-1 (4.34e-2) + & 6.62e-1 (1.98e-2) - & 6.31e-1 (4.40e-2) \\
	 & 10 & 7.15e-1 (6.09e-2) - & 5.97e-1 (2.37e-2) - & 6.09e-1 (2.62e-2) - & 6.13e-1 (1.16e-2) - & 6.19e-1 (8.78e-3) - & 6.43e-1 (2.86e-2) - & 6.35e-1 (2.13e-2) - & \cellbold 5.84e-1 (2.03e-2) \\
	\cline{1-10}
	\multirow{5}{*}{DTLZ5} & 3 & 2.03e-1 (3.90e-2) - & 5.63e-2 (4.49e-2) - & 8.68e-2 (2.19e-2) - & 2.52e-1 (3.16e-2) - & 2.14e-1 (3.31e-2) - & 1.00e-1 (2.86e-2) - & 2.06e-1 (2.98e-2) - & \cellbold 4.09e-2 (1.58e-2) \\
	 & 4 & 1.89e-1 (2.61e-2) - & 1.06e-1 (6.76e-2) + & \cellbold 6.11e-2 (2.28e-2) + & 2.25e-1 (2.87e-2) - & 1.95e-1 (3.11e-2) - & 1.19e-1 (2.52e-2) $\approx$ & 1.89e-1 (2.26e-2) - & 1.10e-1 (2.49e-2) \\
	 & 6 & 1.35e-1 (3.17e-2) - & 1.03e-1 (3.61e-2) + & \cellbold 5.23e-2 (1.20e-2) + & 1.41e-1 (1.59e-2) - & 1.46e-1 (2.54e-2) - & 6.45e-2 (1.99e-2) + & 1.18e-1 (1.95e-2) $\approx$ & 1.20e-1 (2.37e-2) \\
	 & 8 & 8.73e-2 (1.44e-2) - & 7.40e-2 (9.77e-3) - & \cellbold 2.37e-2 (5.86e-3) + & 6.85e-2 (7.06e-3) $\approx$ & 7.43e-2 (1.12e-2) - & 3.45e-2 (9.26e-3) + & 5.50e-2 (5.78e-3) + & 6.89e-2 (1.02e-2) \\
	 & 10 & 2.95e-2 (5.14e-3) - & 1.94e-2 (1.73e-3) $\approx$ & \cellbold 1.25e-2 (2.08e-3) + & 1.94e-2 (2.52e-3) $\approx$ & 1.90e-2 (2.53e-3) $\approx$ & \cellbold 1.22e-2 (2.51e-3) + & 1.39e-2 (1.56e-3) + & 2.01e-2 (3.68e-3) \\
	\cline{1-10}
	\multirow{5}{*}{DTLZ6} & 3 & 6.08e+0 (3.33e-1) - & \cellbold 9.22e-1 (1.87e-1) + & 2.88e+0 (4.35e-1) $\approx$ & 2.21e+0 (8.20e-1) + & 3.48e+0 (5.78e-1) - & 4.30e+0 (6.05e-1) - & 6.06e+0 (3.90e-1) - & 2.77e+0 (9.65e-1) \\
	 & 4 & 5.31e+0 (3.56e-1) + & \cellbold 9.95e-1 (1.77e-1) + & 1.97e+0 (3.26e-1) + & 1.71e+0 (5.97e-1) + & 3.21e+0 (6.65e-1) + & 4.10e+0 (7.11e-1) + & 5.24e+0 (2.20e-1) + & 5.44e+0 (3.04e-1) \\
	 & 6 & 3.76e+0 (2.91e-1) + & \cellbold 6.31e-1 (1.26e-1) + & 1.50e+0 (3.49e-1) + & 1.20e+0 (4.96e-1) + & 2.20e+0 (5.10e-1) + & 2.27e+0 (6.98e-1) + & 3.64e+0 (2.37e-1) + & 4.07e+0 (1.24e-1) \\
	 & 8 & 2.14e+0 (2.80e-1) + & \cellbold 4.35e-1 (1.12e-1) + & \cellbold 5.13e-1 (2.30e-1) + & \cellbold 4.92e-1 (2.18e-1) + & 1.04e+0 (3.77e-1) + & 8.31e-1 (4.07e-1) + & 1.89e+0 (1.76e-1) + & 2.37e+0 (9.13e-2) \\
	 & 10 & 5.31e-1 (2.14e-1) + & 1.44e-1 (5.45e-2) + & \cellbold 6.25e-2 (2.98e-2) + & 2.21e-1 (8.24e-2) + & \cellbold 6.70e-2 (3.34e-2) + & \cellbold 8.50e-2 (9.64e-2) + & 4.82e-1 (1.98e-1) + & 7.01e-1 (3.69e-2) \\
	\cline{1-10}
	\multirow{5}{*}{DTLZ7} & 3 & 4.87e+0 (8.89e-1) - & 1.51e-1 (1.48e-2) + & \cellbold 1.29e-1 (1.90e-2) + & 3.04e-1 (1.04e-1) + & 3.73e+0 (1.30e+0) - & 8.11e-1 (5.20e-1) - & 4.78e+0 (1.13e+0) - & 5.11e-1 (2.17e-1) \\
	 & 4 & 5.66e+0 (1.41e+0) - & \cellbold 3.04e-1 (3.35e-2) + & \cellbold 3.15e-1 (3.18e-2) + & 5.32e-1 (5.53e-2) + & 4.49e+0 (2.12e+0) - & 1.27e+0 (4.78e-1) - & 6.25e+0 (1.65e+0) - & 7.12e-1 (2.63e-1) \\
	 & 6 & 8.57e+0 (2.66e+0) - & 7.48e-1 (3.70e-2) + & \cellbold 6.17e-1 (2.84e-2) + & 9.28e-1 (7.24e-2) + & 4.23e+0 (2.51e+0) - & 2.86e+0 (4.69e-1) - & 8.03e+0 (2.52e+0) - & 1.10e+0 (1.13e-1) \\
	 & 8 & 8.80e+0 (3.16e+0) - & 1.18e+0 (4.96e-2) + & \cellbold 9.63e-1 (6.06e-2) + & 1.12e+0 (4.02e-2) + & 2.79e+0 (1.94e+0) - & 3.53e+0 (5.02e-1) - & 6.35e+0 (2.83e+0) - & 1.77e+0 (2.09e-1) \\
	 & 10 & 3.03e+0 (1.51e+0) - & 1.41e+0 (5.75e-2) + & \cellbold 1.05e+0 (2.92e-2) + & 1.26e+0 (3.21e-2) + & 1.56e+0 (1.09e-1) + & 2.21e+0 (3.91e-1) - & 1.79e+0 (2.79e-1) + & 1.89e+0 (2.11e-1) \\
	\cline{1-10}
	 \multicolumn{2}{c}{+/-/$\approx$} & \multicolumn{1}{c}{10/20/5} & \multicolumn{1}{c}{23/8/4} & \multicolumn{1}{c}{25/5/5} & \multicolumn{1}{c}{19/8/8} & \multicolumn{1}{c}{13/16/6} & \multicolumn{1}{c}{23/11/1} & \multicolumn{1}{c}{16/17/2} & \\
	\bottomrule
	\end{tabular}

	}}
\end{table*}

\begin{table*}
	\caption{
		Statistical results for the IGD metric of 32 runs of each optimizer on the WFG problem set with two objectives. Cells which are among the best performing for that problem up to a 95\% significance level are highlighted.
	}
	\label{tab:high-dimension-results}
    \adjustbox{width=\textwidth + 3cm, center}{
	\centering
	\vspace{3mm}
	\normalsize{
		\begin{tabular}{llllllllll}
	\toprule
	\multicolumn{1}{c}{\textbf{Problem }} &\multicolumn{1}{c}{\textbf{ n }}  &  \multicolumn{1}{c}{\textbf{ NSGA-II }}&  \multicolumn{1}{c}{\textbf{ ParEGO }}&  \multicolumn{1}{c}{\textbf{ K-RVEA }}&  \multicolumn{1}{c}{\textbf{ MOEA/D-EGO }}&  \multicolumn{1}{c}{\textbf{ CPS-MOEA }}&  \multicolumn{1}{c}{\textbf{ REMO }}&  \multicolumn{1}{c}{\textbf{ PC-SAEA }}&  \multicolumn{1}{c}{\textbf{ CRSEA }}\\
	\midrule
	\multirow{3}{*}{WFG1} & 16 & 1.44e+0 (2.41e-1) $\approx$ & 1.32e+0 (4.70e-3) + & 1.35e+0 (4.90e-2) + & 1.38e+0 (2.90e-2) $\approx$ & 1.36e+0 (1.33e-2) $\approx$ & \cellbold 1.30e+0 (5.23e-2) + & 1.42e+0 (1.27e-1) $\approx$ & 1.50e+0 (1.69e-1) \\
	 & 32 & 1.57e+0 (3.33e-1) $\approx$ & \cellbold 1.32e+0 (3.71e-3) + & \cellbold 1.33e+0 (3.50e-2) + & 1.40e+0 (4.26e-2) + & 1.37e+0 (1.22e-2) + & \cellbold 1.32e+0 (4.91e-2) + & 1.58e+0 (2.94e-1) $\approx$ & 1.53e+0 (1.79e-1) \\
	 & 64 & 1.49e+0 (2.52e-1) + & - & \cellbold 1.36e+0 (4.86e-2) + & 1.42e+0 (4.65e-2) + & 1.37e+0 (8.70e-3) + & \cellbold 1.34e+0 (4.97e-2) + & 1.40e+0 (1.21e-1) + & 1.58e+0 (1.73e-1) \\
	\cline{1-10}
	\multirow{3}{*}{WFG2} & 16 & \cellbold 5.14e-1 (1.31e-1) + & 6.32e-1 (3.14e-2) - & \cellbold 4.85e-1 (6.56e-2) + & 5.37e-1 (4.14e-2) + & 5.72e-1 (5.51e-2) $\approx$ & \cellbold 4.87e-1 (6.19e-2) + & 5.47e-1 (6.66e-2) $\approx$ & 5.66e-1 (6.39e-2) \\
	 & 32 & 6.34e-1 (1.53e-1) + & 6.69e-1 (2.12e-2) $\approx$ & 6.56e-1 (2.73e-2) $\approx$ & 5.97e-1 (4.17e-2) + & 6.05e-1 (5.70e-2) + & \cellbold 5.43e-1 (6.10e-2) + & 6.64e-1 (4.92e-2) $\approx$ & 6.58e-1 (4.87e-2) \\
	 & 64 & \cellbold 6.51e-1 (1.22e-1) + & - & 7.16e-1 (4.09e-2) $\approx$ & 6.68e-1 (3.26e-2) + & 6.54e-1 (4.50e-2) + & \cellbold 6.15e-1 (5.01e-2) + & 6.97e-1 (4.31e-2) $\approx$ & 7.09e-1 (2.78e-2) \\
	\cline{1-10}
	\multirow{3}{*}{WFG3} & 16 & 4.00e-1 (5.11e-2) - & 6.04e-1 (2.17e-2) - & 3.04e-1 (6.78e-2) - & 4.90e-1 (2.61e-2) - & 5.21e-1 (3.37e-2) - & 4.27e-1 (4.36e-2) - & 4.31e-1 (4.07e-2) - & \cellbold 2.24e-1 (4.96e-2) \\
	 & 32 & 4.93e-1 (4.00e-2) - & 6.54e-1 (1.53e-2) - & 5.35e-1 (4.02e-2) - & 5.35e-1 (2.31e-2) - & 5.76e-1 (2.55e-2) - & 5.24e-1 (3.69e-2) - & 5.53e-1 (2.22e-2) - & \cellbold 4.61e-1 (4.58e-2) \\
	 & 64 & \cellbold 5.86e-1 (2.48e-2) + & - & 6.33e-1 (1.61e-2) - & 6.38e-1 (1.49e-2) - & 6.16e-1 (2.79e-2) $\approx$ & 5.96e-1 (2.15e-2) + & 6.31e-1 (9.93e-3) - & 6.15e-1 (2.38e-2) \\
	\cline{1-10}
	\multirow{3}{*}{WFG4} & 16 & 2.68e-1 (7.70e-2) - & 3.86e-1 (2.41e-2) - & 3.06e-1 (2.78e-2) - & 3.36e-1 (6.10e-2) - & 3.09e-1 (2.73e-2) - & 2.66e-1 (2.94e-2) - & 3.00e-1 (5.66e-2) - & \cellbold 1.49e-1 (1.45e-2) \\
	 & 32 & 3.36e-1 (6.65e-2) - & 4.75e-1 (1.97e-2) - & 3.78e-1 (2.01e-2) - & 3.85e-1 (5.12e-2) - & 3.63e-1 (1.94e-2) - & 3.24e-1 (4.93e-2) - & 3.45e-1 (3.48e-2) - & \cellbold 2.40e-1 (2.49e-2) \\
	 & 64 & \cellbold 4.01e-1 (9.31e-2) $\approx$ & - & 4.21e-1 (2.72e-2) - & 4.00e-1 (2.09e-2) - & 3.88e-1 (2.22e-2) $\approx$ & \cellbold 3.76e-1 (3.46e-2) $\approx$ & 3.98e-1 (2.38e-2) - & \cellbold 3.83e-1 (2.56e-2) \\
	\cline{1-10}
	\multirow{3}{*}{WFG5} & 16 & 3.66e-1 (4.02e-2) - & 1.09e-1 (1.38e-2) - & 1.78e-1 (2.93e-2) - & 2.20e-1 (4.24e-2) - & 3.30e-1 (2.44e-2) - & 2.84e-1 (4.19e-2) - & 4.37e-1 (3.56e-2) - & \cellbold 9.37e-2 (1.50e-2) \\
	 & 32 & 4.83e-1 (5.60e-2) - & 1.66e-1 (5.15e-2) - & 3.39e-1 (3.24e-2) - & 3.09e-1 (5.78e-2) - & 3.78e-1 (2.42e-2) - & 4.03e-1 (3.84e-2) - & 4.98e-1 (1.95e-2) - & \cellbold 1.17e-1 (1.48e-2) \\
	 & 64 & 5.69e-1 (4.75e-2) - & - & 4.88e-1 (3.11e-2) - & 4.13e-1 (4.95e-2) - & 4.20e-1 (1.95e-2) - & 5.13e-1 (3.44e-2) - & 5.78e-1 (1.89e-2) - & \cellbold 1.79e-1 (2.58e-2) \\
	\cline{1-10}
	\multirow{3}{*}{WFG6} & 16 & 4.11e-1 (5.76e-2) - & \cellbold 2.96e-1 (6.06e-2) $\approx$ & 5.04e-1 (5.19e-2) - & 4.45e-1 (8.56e-2) - & 6.24e-1 (3.75e-2) - & 4.75e-1 (5.08e-2) - & 5.37e-1 (3.64e-2) - & \cellbold 3.26e-1 (8.07e-2) \\
	 & 32 & 5.58e-1 (4.90e-2) - & \cellbold 2.84e-1 (6.22e-2) $\approx$ & 5.81e-1 (4.22e-2) - & 6.20e-1 (8.78e-2) - & 7.23e-1 (3.89e-2) - & 5.72e-1 (4.80e-2) - & 6.42e-1 (3.11e-2) - & \cellbold 3.46e-1 (1.62e-1) \\
	 & 64 & 6.75e-1 (3.76e-2) - & - & 6.93e-1 (2.55e-2) - & 6.89e-1 (5.92e-2) - & 7.51e-1 (2.78e-2) - & 6.72e-1 (3.18e-2) - & 7.15e-1 (3.99e-2) - & \cellbold 4.66e-1 (1.03e-1) \\
	\cline{1-10}
	\multirow{3}{*}{WFG7} & 16 & 3.62e-1 (5.56e-2) - & 4.69e-1 (5.63e-2) - & 4.23e-1 (3.19e-2) - & 4.37e-1 (2.62e-2) - & 4.32e-1 (2.24e-2) - & 3.64e-1 (5.33e-2) - & 3.99e-1 (3.77e-2) - & \cellbold 1.72e-1 (2.94e-2) \\
	 & 32 & 4.51e-1 (5.29e-2) - & 5.83e-1 (2.44e-2) - & 5.28e-1 (1.77e-2) - & 4.93e-1 (3.09e-2) - & 4.85e-1 (1.74e-2) - & 4.48e-1 (3.01e-2) - & 4.71e-1 (2.74e-2) - & \cellbold 3.53e-1 (3.03e-2) \\
	 & 64 & 5.31e-1 (5.64e-2) - & - & 5.76e-1 (9.12e-3) - & 5.43e-1 (1.65e-2) - & 5.18e-1 (2.08e-2) - & \cellbold 5.13e-1 (2.97e-2) $\approx$ & 5.32e-1 (2.52e-2) - & \cellbold 5.05e-1 (2.14e-2) \\
	\cline{1-10}
	\multirow{3}{*}{WFG8} & 16 & 4.75e-1 (4.88e-2) - & 6.46e-1 (4.49e-2) - & 4.32e-1 (3.53e-2) - & 5.79e-1 (2.84e-2) - & 5.74e-1 (4.10e-2) - & 4.65e-1 (4.16e-2) - & 5.63e-1 (4.06e-2) - & \cellbold 3.50e-1 (3.54e-2) \\
	 & 32 & 5.10e-1 (5.30e-2) - & 6.73e-1 (2.88e-2) - & 4.71e-1 (3.25e-2) - & 5.75e-1 (2.38e-2) - & 5.83e-1 (2.93e-2) - & 4.87e-1 (4.13e-2) - & 5.71e-1 (4.74e-2) - & \cellbold 4.31e-1 (3.41e-2) \\
	 & 64 & 5.68e-1 (5.70e-2) - & - & 5.85e-1 (2.68e-2) - & 6.04e-1 (2.51e-2) - & 5.93e-1 (2.85e-2) - & 5.17e-1 (2.93e-2) - & 5.76e-1 (2.54e-2) - & \cellbold 4.86e-1 (3.17e-2) \\
	\cline{1-10}
	\multirow{3}{*}{WFG9} & 16 & 3.90e-1 (7.57e-2) - & 3.61e-1 (6.84e-2) - & 3.84e-1 (8.81e-2) - & 5.27e-1 (6.49e-2) - & 3.96e-1 (5.62e-2) - & 3.66e-1 (1.10e-1) - & 4.18e-1 (7.64e-2) - & \cellbold 2.13e-1 (4.14e-2) \\
	 & 32 & 5.25e-1 (1.01e-1) - & \cellbold 4.17e-1 (7.48e-2) $\approx$ & 6.23e-1 (1.02e-1) - & 6.48e-1 (4.87e-2) - & 4.46e-1 (4.53e-2) - & 5.49e-1 (1.29e-1) - & 6.15e-1 (6.07e-2) - & \cellbold 3.87e-1 (1.11e-1) \\
	 & 64 & 6.59e-1 (6.99e-2) $\approx$ & - & 7.77e-1 (3.87e-2) - & 7.30e-1 (4.86e-2) - & \cellbold 5.48e-1 (8.14e-2) + & 6.83e-1 (8.90e-2) $\approx$ & 7.14e-1 (2.67e-2) - & 6.74e-1 (8.00e-2) \\
	\cline{1-10}
	 \multicolumn{2}{c}{+/-/$\approx$} & \multicolumn{1}{c}{5/18/4} & \multicolumn{1}{c}{2/12/4} & \multicolumn{1}{c}{4/21/2} & \multicolumn{1}{c}{5/21/1} & \multicolumn{1}{c}{5/18/4} & \multicolumn{1}{c}{7/17/3} & \multicolumn{1}{c}{1/21/5} & \\
	\bottomrule
	\end{tabular}

	}}
\end{table*}

Parameters for the new comparison-relationship-surrogate optimizer were selected as follows.
The number of new function evaluations ($\mu$ in Alg.~\ref{alg:crsea}) was set to two and the number of ``model generations'' where the optimizer is run only on the surrogate model ($w_{\text{max}}$) was set to 32.
We discuss more about this choice in a sensitivity study on the parameter in the following subsection.
Two hidden layers of size 64 were used in the feature extraction networks and one hidden layer of size 32 was used for the comparison networks.
The number of epochs in the initial training of the model was 128 with 16 epochs used in every retraining in further generations.
The learning rate was 0.001 at the beginning of each training cycle.
An L2 regularization loss term was used (i.e. through the AdamW optimizer~\cite{loshchilov_decoupled_2019}) with a weight of $\lambda=0.1$.

The first experiment was performed with the popular DTLZ test suite of problems~\cite{deb_scalable_2002} with ten decision variables and three, four, six, eight, and ten objectives.
These results are shown in Tab.~\ref{tab:many-objective-results}.
Each cell lists the mean value of the inverted generational distance of the 32 final approximate Pareto fronts to points from the analytical Pareto front with the standard deviation of the values shown in parentheses.
These values can be sensitive to the number of points used in the reference set and in our work we chose 1000 values evenly distributed along the Pareto front.
Also in each cell is either a plus sign, minus sign, or equals sign indicating whether, for a single problem, the algorithm beats (plus), loses to (minus), or is indistinct from (equals) the new comparison-relationship-surrogate algorithm up to a significance level of 95\%.
Cells are highlighted if they are among the best performing on that problem up to the same significance level.
The performance of the algorithms on DTLZ5 ($m=3$) is also shown in Fig.~\ref{fig:dtlz-performance}.

Although the new algorithm is among the best performing on a handful of problems in the DTLZ test suite, it did not achieve that distinction on the majority of them.
Because the algorithm lacks a diversity preservation mechanism (known to be important for many-objective problems~\cite{cheng_reference_2016}) this was not a surprising result.
We also point out here that the recall of the domination classifier is $r = p^m$ for a problem with $m$ objectives and an underlying comparison model with recall $p$ for each objective.
That is, a true positive prediction of domination requires the correct prediction of every comparison of the objectives.
This points to degraded predictions as the number of objectives grows.
These two properties of the algorithm may explain its performance on the DTLZ test suite.

With a better understanding of the types of problems the comparison-relationship-surrogate optimizer is suited for, we sought out a test with fewer objectives and more decision variables to take advantage of the unique benefits of a neural-network-based surrogate model over those which use a Gaussian process.
To this end, we performed tests on the biobjective WFG test suite \cite{huband_review_2006} with 16, 32, and 64 decision variables.
The number of ``position parameters'' ($k$ in Ref.~\cite{huband_review_2006}) was fixed at two.
These results are shown in Tab.~\ref{tab:high-dimension-results} and show a clear improvement in the quality of solutions generated by the comparison-relationship-surrogate optimizer over the other algorithms studied.
The new algorithm produced among the best solutions seen for 70\% of the problems studied.
For the case of $n=64$, ParEGO was allowed to run on the problems for several days before terminating after not enough progress was made.
An example of the solutions generated for WFG7 ($n=16$) is shown in Fig.~\ref{fig:wfg-combined}b.
The evolution of the quality of the solutions (measured by the IGD metric) during an optimization is visualized for the same example in Fig.~\ref{fig:wfg-combined}a.

\subsection{Sensitivity to Number of Iterations Using Model}
In this section, we evaluate the sensitivity of the CRSEA algorithm to the number of iterations of the evolutionary algorithm run using the model only ($w_{\text{max}}$ in Alg.~\ref{alg:crsea}).
This parameter should be tuned to squeeze the most information out of the model without drifting too far into the regions where it does not have training data.
For a test of how this parameter affects the performance of the algorithm, we have rerun CRSEA on the WFG problem suite with $n=16$ as we vary $w_{\text{max}}$.
We report this data in Tab.~\ref{tab:sensitivity-study}.
All of the settings besides $w_{\text{max}}$ were kept the same as in previous experiments.
Every problem was evaluated 32 times and the Wilcoxon rank sum test used to compare them to the suggested set-point.
These results suggest $w_{\text{max}}=32$ as a reasonable choice for this set of problems.
 
\begin{table*}
	\caption{
		Sensitivity study of the $w_{\text{max}}$ parameter in CRSEA. Statistical results (mean with standard deviation in parenthesis) are shown for the value of the IGD metric of 32 runs using each setting on the WFG (n=16) problems. Cells which are among the best performing for that problem up to a 95\% significance level are highlighted.
	}
	\label{tab:sensitivity-study}
	\adjustbox{width=\textwidth + 3cm, center}{
		\centering
		\vspace{3mm}
		\normalsize{
			\begin{tabular}{lllllll}
    \toprule
    \multicolumn{1}{c}{\textbf{Problem }} &\multicolumn{1}{c}{\textbf{ n }}  &  \multicolumn{1}{c}{\textbf{ CRSEA ($w_{\text{max}}=4$) }}&  \multicolumn{1}{c}{\textbf{ CRSEA ($w_{\text{max}}=8$) }}&  \multicolumn{1}{c}{\textbf{ CRSEA ($w_{\text{max}}=16$) }}&  \multicolumn{1}{c}{\textbf{ CRSEA ($w_{\text{max}}=32$) }}&  \multicolumn{1}{c}{\textbf{ CRSEA ($w_{\text{max}}=64$) }}\\
    \midrule
    WFG1 & 16 & 1.84e+0 (4.43e-2) - & 1.75e+0 (7.98e-2) - & 1.65e+0 (1.54e-1) - & \cellbold 1.47e+0 (1.86e-1) & \cellbold 1.42e+0 (1.43e-1) $\approx$ \\
    \cline{1-7}
    WFG2 & 16 & \cellbold 5.46e-1 (6.70e-2) $\approx$ & \cellbold 5.48e-1 (6.01e-2) $\approx$ & \cellbold 5.63e-1 (6.22e-2) $\approx$ & \cellbold 5.57e-1 (5.33e-2) & 5.90e-1 (5.35e-2) - \\
    \cline{1-7}
    WFG3 & 16 & 3.08e-1 (6.37e-2) - & 2.57e-1 (4.37e-2) - & 2.44e-1 (4.04e-2) $\approx$ & \cellbold 2.32e-1 (4.90e-2) & \cellbold 2.15e-1 (4.22e-2) $\approx$ \\
    \cline{1-7}
    WFG4 & 16 & 1.99e-1 (3.57e-2) - & 1.78e-1 (2.89e-2) - & 1.62e-1 (2.17e-2) - & \cellbold 1.50e-1 (1.37e-2) & \cellbold 1.52e-1 (1.30e-2) $\approx$ \\
    \cline{1-7}
    WFG5 & 16 & 1.81e-1 (2.99e-2) - & 1.13e-1 (2.15e-2) - & 9.52e-2 (7.45e-3) - & \cellbold 9.15e-2 (9.40e-3) & \cellbold 9.04e-2 (6.84e-3) $\approx$ \\
    \cline{1-7}
    WFG6 & 16 & 4.14e-1 (6.30e-2) - & 3.34e-1 (6.17e-2) $\approx$ & 3.24e-1 (6.09e-2) $\approx$ & 3.16e-1 (7.45e-2) & \cellbold 2.80e-1 (6.75e-2) + \\
    \cline{1-7}
    WFG7 & 16 & 2.53e-1 (4.37e-2) - & 2.03e-1 (4.59e-2) - & \cellbold 1.88e-1 (3.82e-2) $\approx$ & \cellbold 1.71e-1 (3.28e-2) & \cellbold 1.76e-1 (3.86e-2) $\approx$ \\
    \cline{1-7}
    WFG8 & 16 & 3.91e-1 (3.02e-2) - & 3.65e-1 (3.46e-2) - & 3.61e-1 (3.65e-2) - & \cellbold 3.47e-1 (3.06e-2) & \cellbold 3.51e-1 (3.57e-2) $\approx$ \\
    \cline{1-7}
    WFG9 & 16 & 2.92e-1 (9.69e-2) - & 2.46e-1 (6.59e-2) - & \cellbold 2.24e-1 (5.34e-2) $\approx$ & \cellbold 2.07e-1 (6.07e-2) & \cellbold 2.25e-1 (4.68e-2) $\approx$ \\
    \cline{1-7}
     \multicolumn{2}{c}{+/-/$\approx$} & \multicolumn{1}{c}{0/8/1} & \multicolumn{1}{c}{0/7/2} & \multicolumn{1}{c}{0/4/5} &  & \multicolumn{1}{c}{1/1/7}\\
    \bottomrule
\end{tabular}
	}}
\end{table*}

\section{Performance on a Real-World Problem}
\label{sec:performance-real-world-problem}
As another benchmark of the comparison-relationship-surrogate optimizer, we chose to evaluate it on a practical medium-scale problem from the field of accelerator physics.
This problem is a photoinjector optimization for ultrafast electron diffraction applications~\cite{pierce_low_2020} with 13 decision variables.
Due to the high computational cost of the test (20 to 30 minutes per function evaluation) and limits on the programming language the simulation tool and its interfaces are written in, it is restricted to a comparison between the new comparison-based evolutionary algorithm and a single conventional optimizer (NSGA-II here).

The problem setup is identical to the DC gun ultrafast electron diffraction optimization problem described in Ref.~\cite{pierce_low_2020}, with the following changes.
First, constraints are not used in this test to make it compatible with the new optimization algorithm.
Because these constraints protect the optimization against parts of decision space that may include beam loss, we also reduce the range of the following decision variables to avoid that condition.
The two solenoid current settings are constrained to the ranges \SIrange{2}{3}{\ampere}  and \SIrange{1.5}{2.5}{\ampere} respectively.
The buncher's peak voltage is also restricted to a range of \SIrange{10}{30}{\kV}.
With these changes, we found good transmission of the beam across the space of decision variables.
We also manually confirmed that the non-dominated solutions returned from the optimization algorithm had no beam loss.
The cathode's mean transverse energy was set to \SI{100}{\milli\electronvolt} and the number of macroparticles used was 10000.
The two objectives used in the optimization are: transverse emittance of the beam at the sample location (minimize) and the delivered bunch charge (maximize).

\begin{figure}[h!]
	\centering
	\includegraphics{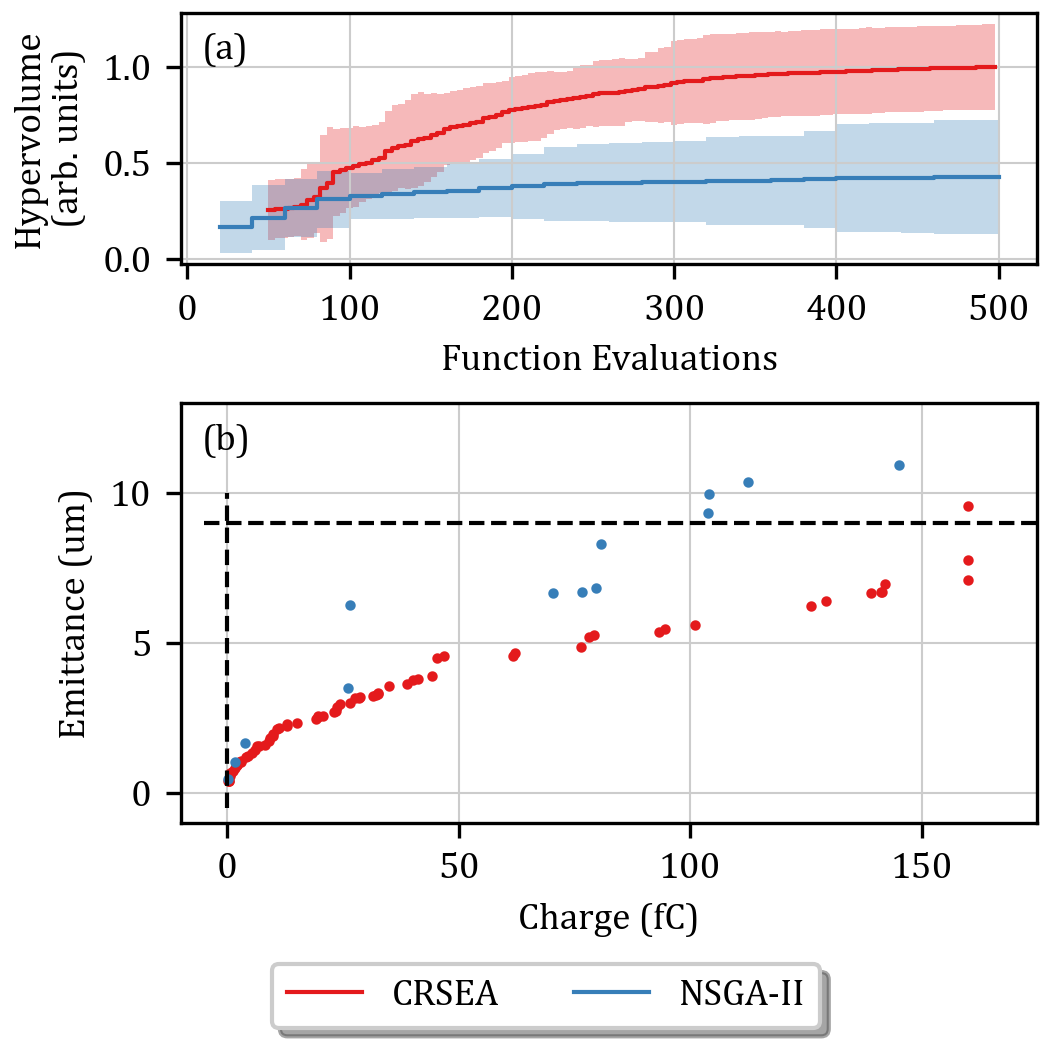}
	\caption{Results from the ultrafast electron diffraction photoinjector optimization problem: (a) Mean hypervolume metric for non-dominated solutions with the 95\% prediction interval shown as the shaded regions; (b) The solutions after 500 function evaluations from the optimization with median value of the hypervolume for each algorithm. The reference point is shown on this plot as the dashed lines.}
	\label{fig:photoinjector-combined}
\end{figure}

Each optimization was run 8 times with a budget of 500 function evaluations.
The same algorithm settings were used as in the biobjective experiments in Sec.~\ref{sec:empirical-studies} except for starting with an initial population size of 50.
The hypervolume of the approximate fronts was calculated using a reference point of (\SI{0}{\femto\coulomb}, \SI{9}{\nano\meter}).
The mean hypervolume and a band indicating the 95\% prediction interval are shown versus the number of function evaluations in Fig.~\ref{fig:photoinjector-combined}a.
Examples of the non-dominated solutions returned from the optimization algorithms are also shown in Fig.~\ref{fig:photoinjector-combined}b.

\section{Conclusion}
\label{sec:conclusion}
In this work, we investigated the use of a comparison-relationship-surrogate model to improve the quality of solutions achievable with multi-objective genetic algorithms on a fixed budget of function evaluations.
This type of model, given two individuals, will predict which of them has the smaller value of each objective function.
Domination relationships can then be inferred after a ``cleaning'' procedure allowing the model to be integrated into a non-dominated sorting genetic algorithm.
Its performance was then compared to that of a selection of other algorithms on the DTLZ and WFG test suites as well as a real-world problem from the field of accelerator physics.

Our numerical experiments found that the comparison-relationship-surrogate optimizer produced among the best solutions of all the algorithms studied for many \emph{medium-scale, biobjective} problems (specifically WFG3-WFG9 in Tab.~\ref{tab:high-dimension-results}).
Tests on the real-world optimization problem also showed the ability of the algorithm to find a larger solution hypervolume than the tested conventional genetic optimizer (NSGA-II).
Results were not as promising on the DTLZ test suite where solution diversity may have limited performance.

Finding diversity preservation methods which are compatible with predictions from the comparison-relationship-surrogate model will be an important line of research coming out of this work.
Inclusion of constraints using a model of feasibility is also a natural extension to the work as it can be naturally added to the prediction of domination in the same way as in Ref.~\cite{deb_fast_2002}.
With continued work in this area, these algorithms could help to improve the results available to the scientific and industrial users of optimization algorithms that face expensive multi-objective problems on a limited budget.

\section{CRediT Authorship Contribution Statement}
{\bf Christopher M. Pierce}: Conceptualization, Investigation, Software, Visualization, Writing – original draft, Writing – review \& editing.
{\bf Young-Kee Kim}: Funding acquisition, Supervision, Writing – review \& editing.
{\bf Ivan Bazarov}: Funding acquisition, Supervision, Writing – review \& editing.

\section{Declaration of Competing Interest}
The authors declare that they have no known competing financial interests or personal relationships that could have appeared to influence the work reported in this paper.

\section{Acknowledgments}
This work was supported by the U.S. National Science Foundation under Award PHY-1549132, the Center for Bright Beams.

\bibliographystyle{elsarticle-num} 
\bibliography{comparison-relationship-surrogate-evolutionary-algorithm.bib}

\end{document}